\documentclass[10pt,twocolumn,letterpaper]{article}

\usepackage{wacv}
\usepackage{times}
\usepackage{epsfig}
\usepackage{graphicx}
\usepackage{amsmath}
\usepackage{amssymb}

\usepackage{amssymb}% http://ctan.org/pkg/amssymb
\usepackage{kotex} % korean
\usepackage{tabularx} % in the preamble
\usepackage{subfigure}
\usepackage{verbatim}  % comment
\usepackage{adjustbox}
\usepackage{cite}
\usepackage{url}
\usepackage{amsmath, soul}
\usepackage[ruled]{algorithm2e}

\wacvfinalcopy % *** Uncomment this line for the final submission

\ifwacvfinal
\def\assignedStartPage{1} % *** Enter the assigned starting page number (instead of 9876)
\fi

\usepackage[pagebackref=true,breaklinks=true,colorlinks,bookmarks=false]{hyperref}

\ifwacvfinal
\else
\fi

\begin{document}

\definecolor{brown}{rgb}{0.59, 0.29, 0.0}
\newcommand\by[1]{{\textcolor{red}{#1}}}
\newcommand\erase[1]{{\textcolor{brown}{#1}}}

%%%%%%%%% TITLE
\title{TricubeNet: 2D Kernel-Based Object Representation for Weakly-Occluded Oriented Object Detection}

\newcommand{\customfootnotetext}[2]{{% Group to localize change to footnote
\renewcommand{\thefootnote}{#1}% Update footnote counter representation
\footnotetext[0]{#2}}}% Print footnote text

\author{

Beomyoung Kim$^{1\textsuperscript{$\dagger$}}$\hspace{1.5em}Janghyeon Lee$^{2\textsuperscript{$\dagger$}}$\hspace{1.5em}Sihaeng Lee$^{2\textsuperscript{$\dagger$}}$\hspace{1.5em}Doyeon Kim$^{3}$\hspace{1.5em}Junmo Kim$^{3}$\\
{NAVER CLOVA$^1$\hspace{3em}LG AI Research$^2$\hspace{3em}KAIST$^3$}\\
%{\tt\small beomyoung.kim@navercorp.com, \{janghyeon.lee, sihaeng.lee\}@lgresearch.ai,}\\
%{\tt\small \{doyeon\_kim, junmo.kim\}@kaist.ac.kr}
}

\maketitle

\customfootnotetext{$\dagger$}{This work was done when the author worked at KAIST.}

%\thispagestyle{empty}
%%%%%%%%% ABSTRACT
\begin{abstract}
    We present a novel approach for oriented object detection, named TricubeNet, which localizes oriented objects using visual cues ($i.e.,$ heatmap) instead of oriented box offsets regression.
    We represent each object as a 2D Tricube kernel and extract bounding boxes using simple image-processing algorithms.
    Our approach is able to (1) obtain well-arranged boxes from visual cues, (2) solve the angle discontinuity problem, and (3) can save computational complexity due to our anchor-free modeling.
    To further boost the performance, we propose some effective techniques for size-invariant loss, reducing false detections, extracting rotation-invariant features, and heatmap refinement.
    To demonstrate the effectiveness of our TricubeNet, we experiment on various tasks for weakly-occluded oriented object detection: detection in an aerial image, densely packed object image, and text image.
    The extensive experimental results show that our TricubeNet is quite effective for oriented object detection. Code is available at \url{https://github.com/qjadud1994/TricubeNet}.
\end{abstract}

%%%%%%%%% BODY TEXT

\section{Introduction}

Object detection is one of the fundamental computer vision tasks, and deep learning-based methods~\cite{ren2015faster, cai2018cascade, tan2020efficientdet} have shown remarkable performance.
However, existing detectors often focus on detecting a horizontal bounding box that is not sufficient in some cases.
First, for densely arranged oriented objects, an intersection-over-union (IOU) between adjacent horizontal bounding boxes tends to be large, and some of these boxes will be filtered out by non-maximum suppression (NMS).
Second, since the horizontal bounding box can contain many redundant areas, it is not suitable for real-world applications that require tighter and more accurate boxes, such as aerial images and scene text images.
To detect the object in a more accurate form, oriented object detection has attracted much attention recently.

%-------------------------------------------------------------------------
\begin{figure}[t]
    \centering
    \includegraphics[width=\linewidth]{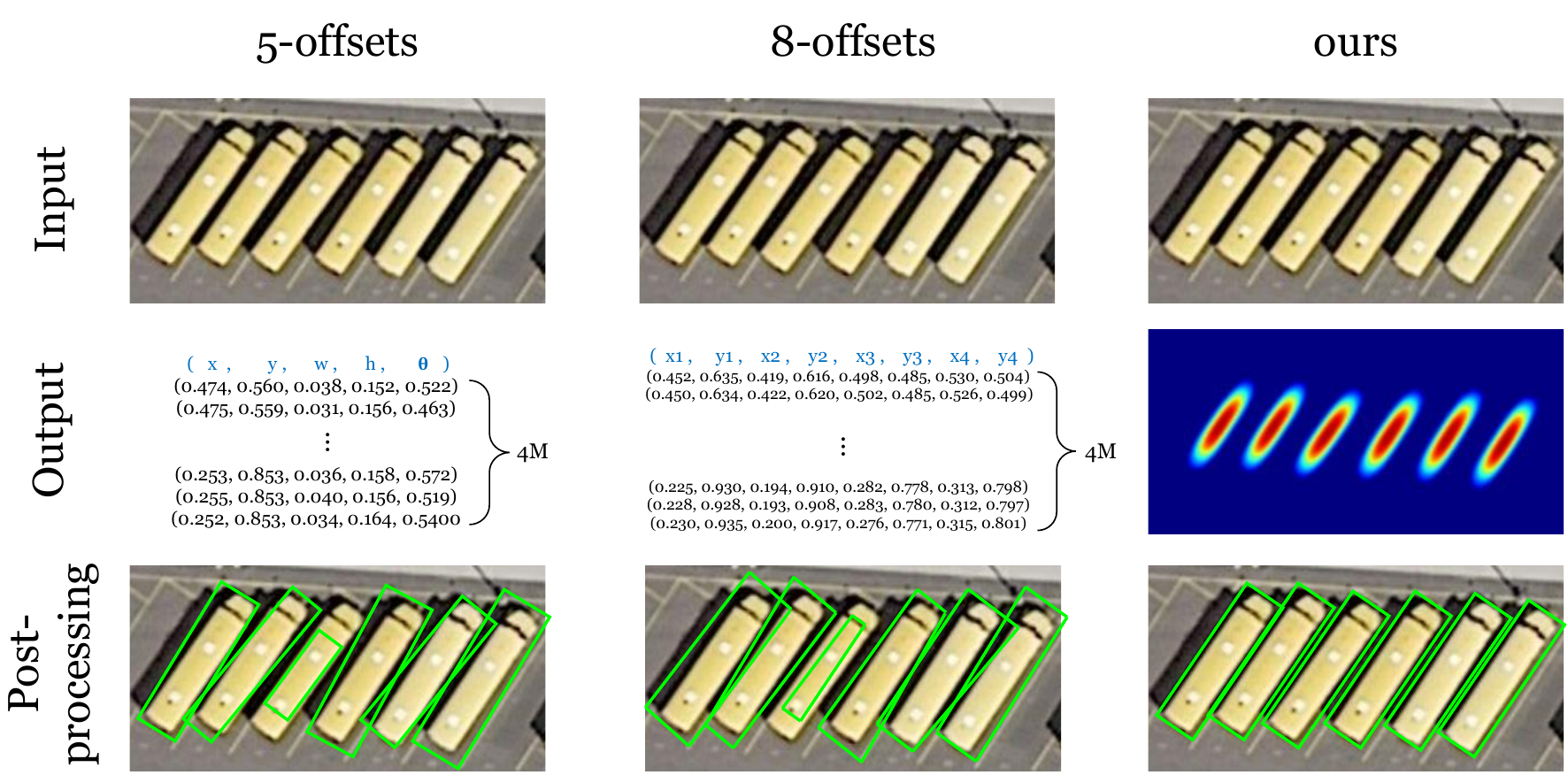}
    \caption{ Existing anchor-based methods that regress five or eight offsets require a huge number of anchor boxes (about 4M) and infer a bit scatty boxes.
    We represent each object as a 2D Tricube kernel and extract bounding boxes using simple image-processing algorithms.
    Our approach, named TricubeNet, does not require the anchor boxes and can obtain well-arranged boxes.
    }
    \label{fig:concept}
    \vspace{-4mm}
\end{figure}
%-------------------------------------------------------------------------

Most oriented object detectors~\cite{ma2018arbitrary, ding2019learning, xu2020gliding} adopt Faster R-CNN~\cite{ren2015faster} or RetinaNet~\cite{lin2017focal} as their baseline model and additionally infer an angle of the object.
They adopt anchor augmentation strategy and regress oriented box offsets in form of 5-offsets ($x$, $y$, $w$, $h$, $\theta$) or 8-offsets ($x1$, $y1$, $x2$, $y2$, $x3$, $y3$, $x4$, $y4$).
They have reigned on the throne with state-of-the-art performance, however, some limitations remain.
(1) Regressing the box offsets might have trouble in obtaining well-arranged boxes of densely arranged oriented objects as in Figure \ref{fig:concept};
(2) Regressing the angle offset causes angle discontinuity problem; the angle discontinuity on the boundary leads to the loss fluctuation during training;
(3) They require huge computational complexity due to the anchor augmentation and a heavy IoU calculation for the oriented box.
For example, when they define anchor boxes with three scales, five aspect ratios, and six angles and adopt FPN~\cite{lin2017feature} architecture with $800 \times 800$ input resolution, they require about 4M total anchor boxes ($3 \times 5 \times 6 = 90$ anchor boxes per a pixel location).
Some might argue that anchor-free approaches such as~\cite{zhou2019objects, pan2020dynamic} can reduce the computational complexity, however, they also suffer from the angle discontinuity problem due to their angle regression.

In this paper, we introduce a novel approach for the oriented object detection, named \textbf{TricubeNet}.
We localize oriented objects using \textit{visual cues} ($i.e., $ heatmap) instead of the box offset regression.
As shown in Figure \ref{fig:concept}, we represent each object as a 2D Tricube kernel whose shape visually describes the width, height, and angle of the object, and then extract bounding boxes using simple image-processing algorithms.
Our approach is able to (1) obtain well-arranged oriented boxes from visual cues of arranged objects as shown in Figure \ref{fig:concept}, (2) solve the angle discontinuity problem by taking away the angle regression, (3) save computational complexity due to our anchor-free modeling, and (4) is a simple one-stage anchor-free detector.

Furthermore, for the competitive result, we should handle some challenging factors of the oriented object detection: various shapes and sizes of objects, densely arranged objects, a huge number of objects, false detections, and complexity of the background.
To deal with these challenging factors, we propose some techniques.
The first is a \textit{Size-Weight Mask} (SWM).
The pixel-wise mean squared error (MSE) loss causes a size-imbalance problem, that is, a small object tends to be given small loss, weakening the detection of small objects.
To make a size-invariant loss function, we propose the SWM.
Second, to give a balanced loss between foreground and background pixels and reduce false-positive detections at once, we introduce a \textit{False-Positive Example Mining} (FPEM) technique.
Third, we propose a \textit{Multi-Angle Convolution} (MAC) module to extract a rotation-invariant feature for the oriented object.
Last, we design a repetitive refinement stage to refine the output heatmap and call this technique \textit{heatmap cascade refinement}.

We verify the effectiveness of the TricubeNet in various tasks: oriented object detection in the aerial image (DOTA~\cite{xia2018dota}), the densely packed object image (SKU110K-R~\cite{goldman2019precise, pan2020dynamic}), and the scene text image (MSRA-TD500~\cite{yao2012detecting}, ICDAR 2015~\cite{karatzas2015icdar}).
We target detection of weakly-occluded oriented objects and choose the above highly practical tasks in real-world applications.
The experimental results show that TricubeNet is quite effective to detect the oriented object with a simple anchor-free one-stage process.

In summary, our contributions are as follows:
\begin{itemize}
    \item We propose a novel oriented object detector, TricubeNet, which localizes oriented objects using visual cues ($i.e.,$ heatmap) instead of the box offset regression.
    \item Our approach can obtain well-arranged oriented boxes, solve the angle discontinuity problem, and save computational complexity by eliminating anchor boxes.
    \item We propose some techniques ($i.e.,$ SWM, FPEM, MAC, and heatmap cascade refinement) to properly detect the oriented object and boost the performance.
    \item We verify the effectiveness of our TricubeNet from extensive experimental results on various tasks.
\end{itemize}

%------------------------------------------------------------------------
\begin{figure*}[t]
    \centering
    \includegraphics[width=\linewidth]{ 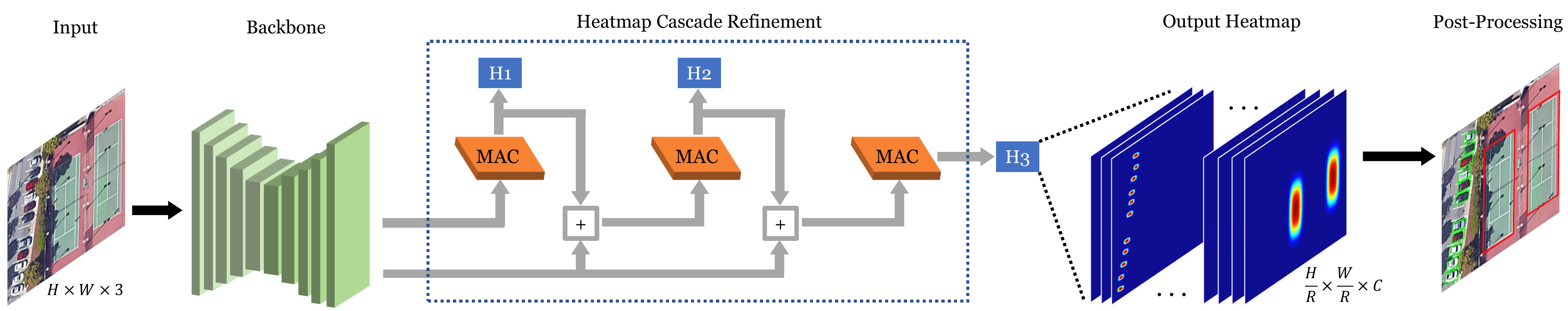}
    \caption{Overview of TricubeNet. It produces one channel heatmap per category where each oriented object is represented as a 2D Tricube kernel. The backbone network consists of a fully convolutional encoder--decoder architecture. $H$ and $W$ are the height and width of the image, respectively; $C$ is the number of categories; $R$ is the downsampling rate. From the heatmap cascade refinement, we progressively refine the output heatmap (H) and extract bounding boxes from the lastly refined heatmap (H$_{3}$) using simple image-processing algorithms.}
    \label{fig:architecture}
\end{figure*}
%-------------------------------------------------------------------------

%-------------------------------------------------------------------------
\begin{figure}[t]
    \centering
    \subfigure[]{           
        \includegraphics[width=0.2\linewidth]{ 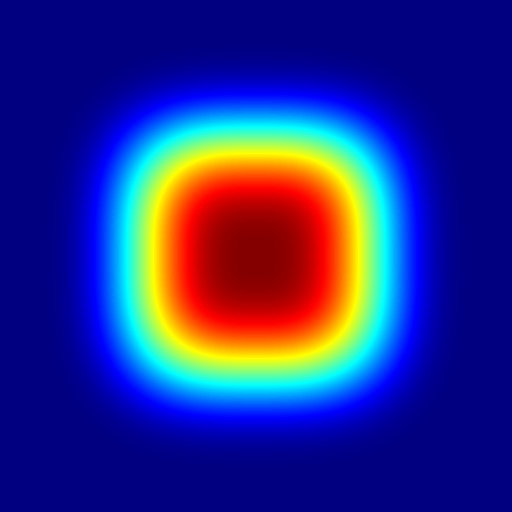} 
        \label{fig:tricube}        
    }
    \subfigure[]{           
        \includegraphics[width=0.2\linewidth]{ 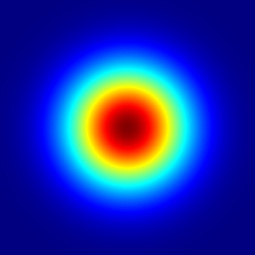} 
        \label{fig:gaussian}          
    }
    \subfigure[]{           
        \includegraphics[width=0.2\linewidth]{ 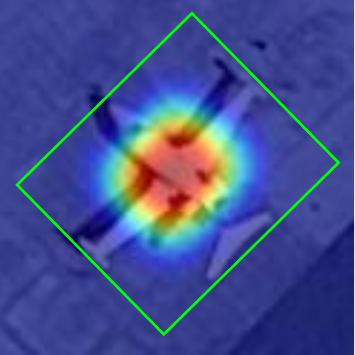} 
        \label{fig:object_as_tricube}        
    }
    \subfigure[]{           
        \includegraphics[width=0.2\linewidth]{ 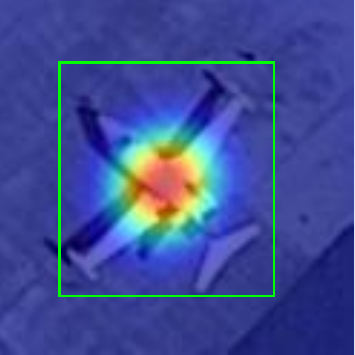} 
        \label{fig:object_as_gaussian}          
    }
    \caption{ (a): 2D Tricube kernel, (b): 2D Gaussian kernel, (c): object as 2D Tricube kernel, and (d): object as 2D Gaussian kernel.}
    \label{result_kernels}
\end{figure}
%-------------------------------------------------------------------------

\section{Related Work}

\textbf{Two-stage object detectors} consist of two processes: extract object region candidates and crop the region of interest (ROI) of each object and predict the class and bounding box offsets of the object. R-CNN~\cite{girshick2014rich} uses a selective search~\cite{uijlings2013selective} method to generate the bounding box candidates and feeds them to the classifier. SPP~\cite{he2015spatial} and Fast R-CNN~\cite{girshick2015fast} crop the ROIs from feature maps and feed them to the classifier. Faster R-CNN~\cite{ren2015faster} generates bounding box candidates from a region proposal network (RPN), which allows training in an end-to-end manner.
Recently, Cascade R-CNN~\cite{cai2018cascade} designs iterative refinement steps for the high-quality bounding box prediction.
Although the two-stage object detectors achieve state-of-the-art performance, they require a high computational complexity.

\noindent \textbf{One-stage object detectors} classify and regress the bounding boxes at once using anchor boxes.
SSD~\cite{liu2016ssd} densely produces the bounding boxes from multi-level feature maps using various sizes of anchor boxes and removes overlapping bounding boxes using NMS post-processing. 
RetinaNet~\cite{lin2017focal} uses a focal loss to alleviate the class-imbalance problem between positive and negative anchor boxes.

\noindent \textbf{Anchor-free object detectors} have recently been proposed and eliminate the anchor box in the network design.
CornerNet~\cite{law2018cornernet} is a keypoint-based anchor-free approach that represents objects as pairs of corner keypoints and groups them. 
CenterNet~\cite{zhou2019objects} represents objects as center points with width and height regression.
The anchor-free detectors are simple and efficient but report a lower performance than the two-stage detectors.

\noindent \textbf{Oriented object detectors} often adopt the object detectors and additionally regress the angle for oriented objects.
Adopting the Faster-RCNN~\cite{ren2015faster} as a baseline detector, RRPN~\cite{ma2018arbitrary} exploits rotated anchor boxes and changes the IoU calculation for rotated boxes, RoI Transformer~\cite{ding2019learning} extracts geometry-robust pooled features, and Gliding vertex~\cite{xu2020gliding} employs a simple object representation method which glides the vertex of the
horizontal bounding box on each corresponding side.
Adopting the RetinaNet~\cite{lin2017focal} as a baseline detector, RSDet~\cite{qian2019learning} proposes an eight-parameter regression model using a rotation sensitivity error (RSE) for handling the angle discontinuity problem and R$^3$Det~\cite{yang2019r3det} is a fast one-stage detector with a feature refinement module to handle the feature misalignment problem.
SCRDet~\cite{yang2019scrdet} proposes an IoU loss to alleviate the angle discontinuity problem.
These approaches adopt an anchor augmentation strategy with a huge amount of box candidates to obtain oriented boxes.
To reduce computational complexity by eliminating anchor box, DRN~\cite{pan2020dynamic} adopts CenterNet~\cite{zhou2019objects} as their baseline detector taking anchor-free modeling and additionally regresses the angle offset with a proposed feature selection module and dynamic refinement head to dynamically refine the network.
Note that DRN represents objects as center points and obtains other box offsets using regression, suffering from the angle discontinuity problem.
Meanwhile, we represent the whole object region as a 2D Tricube kernel and obtain a bounding box using image-processing algorithms, detouring the angle discontinuity problem. 

The text detectors widely have adopted a segmentation-based anchor-free approach by utilizing the fact that there is no occlusion in the text.
PixelLink~\cite{deng2018pixellink} performs the instance segmentation using a pixel-wise eight-neighbors link prediction.
CRAFT~\cite{baek2019character} is a character-level detection approach instead of word-level detection by representing each character as a 2D Gaussian kernel.

\section{Method}

\subsection{2D Tricube kernel} \label{method:kernel}
Recent key-point detection approaches~\cite{law2018cornernet, zhou2019objects, newell2016stacked} represent each key-point as a 2D Gaussian kernel and show remarkable performances.
Motivated by these approaches, we assume that a 2D kernel function is a good candidate for representing the key features and is an easy-to-learn form for deep neural networks. 
For the oriented object detection, however, the 2D Gaussian kernel is not a suitable choice because its circular form often fails to represent the angle of an object.
For example, in Figure \ref{fig:object_as_gaussian}, the 2D Gaussian kernel for the rotated square-form object cannot represent the angle of the object.
For suitable modeling for the oriented object detection, we choose a 2D Tricube kernel as our object representation method.
Unlike the 2D Gaussian kernel that has a circular-form distribution, the 2D Tricube kernel has a rectangular-form distribution as in Figure \ref{fig:tricube} and is defined as $(1-|x|^3)^\gamma\cdot(1-|y|^3)^\gamma$ where $\gamma$ is set to 7 in all experiments.
The 2D Tricube kernel can properly represent the angle of the object as shown in Figure \ref{fig:object_as_tricube}.

%-------------------------------------------------------------------------

\begin{figure}[t]
    \centering
    \includegraphics[width=\linewidth]{ 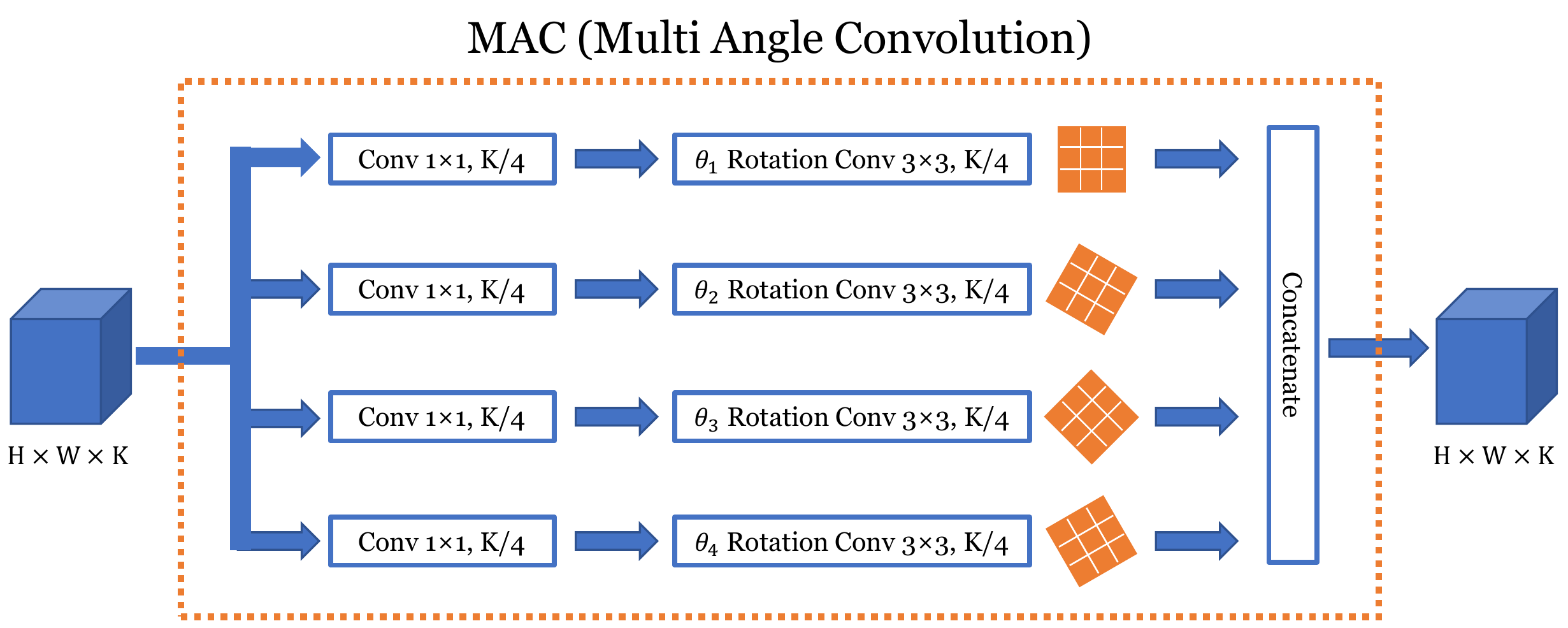}
    \caption{Illustration of multi-angle convolution (MAC) module which extracts rotation-invariant features. }
    \label{fig:mac}
\end{figure}

%-------------------------------------------------------------------------

\subsection{Architecture} \label{architecture}
As illustrated in Figure \ref{fig:architecture}, TricubeNet consists of a fully convolutional encoder--decoder architecture.
Let $I \in \mathbb{R}^{H \times W \times 3}$ be an input image, where $H$ is height and $W$ is width. 
The network predicts the output heatmap $H \in \mathbb{R}^{\frac{H}{R} \times \frac{W}{R} \times C}$, where $R$ is the downsampling rate and $C$ is the number of categories.
We adopt Hourglass-104~\cite{newell2016stacked} network as our backbone network.
We apply a bilinear upsampling layer on the output heatmap for higher pixel precision; the downsampling rate $R$ is set to 2 in all experiments. 

\noindent \textbf{Multi-Angle Convolution (MAC) Module.}
In our heatmap-based detection framework, extracting rotation-invariant features is important to detect orient objects.
However, a convolutional neural network (CNN) has a limitation in extracting a rotation-invariant feature.
To this end, we design a multi-angle convolution (MAC) module that is light and effective to extract the rotation-invariant features.
The MAC module is illustrated in Figure \ref{fig:mac}.

First, we define $n$ kinds of angles, $\theta_{i}$, $i \in (1, \dots, n)$, where the range is [0, $\frac{\pi}{2}$).
Next, when the input feature map $X$ consists of $K$ channels, we divide the channel of the $X$ by $n$ by applying a convolutional layer with a kernel size of 1 and channel of $\frac{K}{n}$.
We denote the divided feature map as $\mathcal{F}_{i}$.
Then, we apply a rotation convolutional layer with an angle of $\theta_{i}$ and kernel size 3 on $\mathcal{F}_{i}$.
Last, we concatenate the outputs of the rotation convolution with $n$ different angles applied.
Formally, the output of the MAC module $\hat{X}$, which is the rotation invariant feature map, is defined as:
\begin{equation}
    \hat{X} = concat([ RConv(\mathcal{F}_{i}, \theta_{i}), \dots, RConv(\mathcal{F}_{n}, \theta_{n})  ]),
\end{equation}
where $concat$ is the concatenate operation and $RConv$ is the rotation convolutional layer with an angle of $\theta$.
In all experiments, we set $n=4$ and $(\theta_{1}, \theta_{2}, \theta_{3}, \theta_{4})=(0, \frac{\pi}{6}, \frac{\pi}{4}, \frac{\pi}{3})$.

A recent study, DRN~\cite{pan2020dynamic}, proposed a rotation convolutional layer, which rotates the kernel of CNN to extract the rotation-invariant feature.
However, rotating the kernel of CNN requires heavy computations and additional parameters.
For the efficient implementation of the rotation convolutional layer, we approximate the rotating of the kernel by rotating feature maps and then applying a normal CNN.

The detailed process of our rotation convolutional layer is described in Algorithm \ref{alg:rotation_conv}.
First, we re-scale the feature map with a scale factor of $1/(\sin{\theta} + \cos{\theta})$ to keep the original shape after rotating and then rotate the feature map by $\theta$.
Then, we apply a normal convolutional layer with kernel size 3 on the rotated feature map and return the output feature map back to the original angle and scale.

\begin{algorithm}[t]
    \SetKwInOut{Input}{Input}
        \Input{
            Feature maps $X \in \mathbb{R}^{H \times W \times K}$, \\ rotation angle $\theta$
        }
    \SetKwInOut{Output}{Output}
        \Output{
            Rotation-invariant feature maps : $\hat{X} \in \mathbb{R}^{H \times W \times K}$
        }
    {
        $\hat{X}$ $\gets$ Rescale(X, $1 / (\sin{\theta} + \cos{\theta})$) \\
        $\hat{X}$ $\gets$ Rotate($\hat{X}$, $\theta$) \\
        $\hat{X}$ $\gets$ Conv3$\times$3($\hat{X}$) \\
        $\hat{X}$ $\gets$ Rotate($\hat{X}$, $2\pi - \theta$) \\
        $\hat{X}$ $\gets$ Rescale($\hat{X}$, $\sin{\theta} + \cos{\theta}$) \\
        
        \textbf{return} $\hat{X}$
    }
\caption{Rotation Convolution}
\label{alg:rotation_conv}
\end{algorithm}

\noindent \textbf{Heatmap Cascade Refinement.}
The two-stage detectors, such as Cascade R-CNN~\cite{cai2018cascade}, take advantage of a progressive cascade refinement of the predicted bounding box; it repeatably crops the region of the predicted bounding box and predicts refined bounding box.
In contrast, the cascade refinement step is less studied on the heatmap-based anchor-free approaches.
Here, in our study, we propose a heatmap cascade refinement, which progressively refines the output heatmap in a pixel-wise manner, as illustrated in Figure \ref{fig:architecture}.
Specifically, when the last feature map our backbone network is denoted as $X$, the $r$-th refined feature map $Y_{r}$ and $r$-th refined heatmap $H_{r}$ are defined as:

\begin{equation}
  \begin{aligned}
    Y_{1} &= MAC(X), \; H_{1} = fc(Y_{1}),\\
    Y_{2} &= MAC(X + Y_{1}),  \; H_{2} = fc(Y_{2}),\\
    \vdots \\
    Y_{r} &= MAC(X + Y_{r-1}),  \; H_{r} = fc(Y_{r}), 
  \end{aligned}
\end{equation}
where $MAC$ is the multi-angle convolution module and $fc$ is a convolutional layer with a kernel size of 1 and a channel of $C$.
Through the heatmap cascade refinement step, we progressively refine the output heatmap by taking the rotation-invariant feature map of the previous refinement step.

%-------------------------------------------------------------------------

\begin{figure}[t]
    \centering
    \includegraphics[width=\linewidth]{ 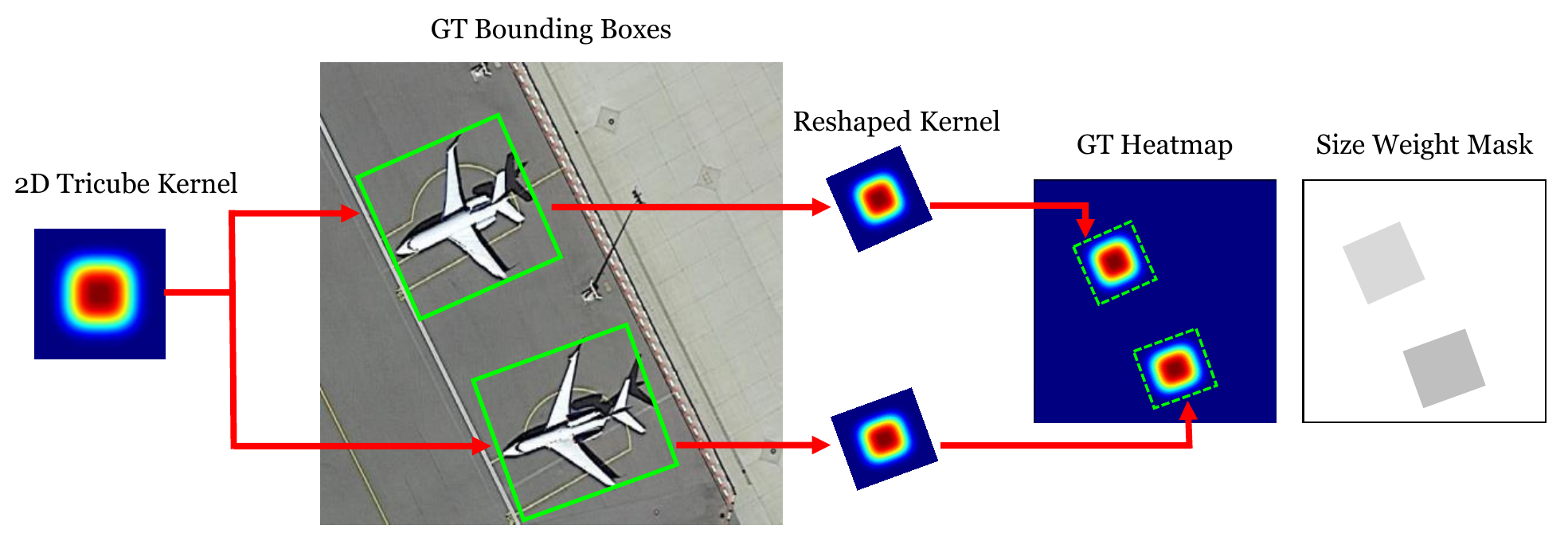}
    \caption{Illustration of the ground truth (GT) heatmap and size-weight mask generation procedure. }
    \label{fig_box2map}
    %\vspace{-5mm}
\end{figure}

%-------------------------------------------------------------------------

\subsection{From bounding boxes to heatmap} \label{box2map}
Here, we describe how to generate a ground truth heatmap.
The whole generation process is illustrated in Figure \ref{fig_box2map}.
First, we create a square 2D Tricube kernel and normalize it to a value between zero to one.
Then, we reshape the kernel to the same size as the ground truth oriented bounding box.
Last, we insert this reshaped kernel into the heatmap.
If two Tricube kernels overlap, we take an element-wise maximum operation.
The generated ground truth heatmap is denoted as $\hat{H} \in \mathbb{R}^{\frac{H}{R} \times \frac{W}{R} \times C}$.

\subsection{Objective} \label{objective}
We optimize the TricubeNet using pixel-wise mean-squared error (MSE) objective function between $H$ and $\hat{H}$
However, giving the pixel-wise MSE loss function has two problems.
First is a size-imbalance problem. 
Since we assign the 2D Tricube kernel according to the size of each object, large objects tend to have large losses and small objects tend to have little losses; this weakens the detection of small objects.
Second is a class-imbalance problem. 
Most of the pixels in the heatmap are background pixels, therefore, it interferes with focusing on the foreground pixels.
To alleviate the above problems, we propose a size-weight mask (SWM) and false-positive example mining (FPEM).

%\vfill
\noindent
\textbf{Size-Weight Mask (SWM).}
For the size-invariant loss function, we weight the MSE loss according to the size of the object using the size-weight mask (SWM); a large object is given a low weight and a small object is given a high weight.
The SWM is denoted as $M \in \mathbb{R}^{\frac{H}{R} \times \frac{W}{R} \times C}$.
For each object, foreground pixels in the SWM contain a weight that is inversely proportional to the size of the object. 
Also, background pixels contain a weight of 1.
This mask is generated in the same manner as the ground truth heatmap (Section~\ref{box2map}) and illustrated in Figure \ref{fig_box2map}.
When we denote the size of the $i$-th object as ${S_i}$, the set of pixels of $i$-th object as $\mathcal{P}_{i}^{obj}$, and the number of objects in an image as $N$, the SWM for $i$-th object, $M_{i}$, is defined as:
\begin{equation}
S = \sum_{i=1}^{N}{S_i}, \quad M_{i}(p) = \begin{cases} \frac{S}{S_i \times N} & p \in \mathcal{P}_{i}^{obj} \\ 1 & \text{otherwise.} \end{cases}
\end{equation}

\noindent
\textbf{False-Positive Example Mining (FPEM).}
To alleviate the class-imbalance problem, we adopt Online Hard Example Mining (OHEM)~\cite{shrivastava2016training} that is an effective sampling method to make more focus on the foreground objects.
However, focusing too much on the foreground objects yields a lot of false-positive detections.
For balanced sampling while reducing false-positives, we propose false-positive exampling mining (FPEM).
Specifically, we extract the positive and false-positive pixels during training in an online manner:
\begin{equation}
    \begin{aligned}
        \mathcal{P}^{pos} &=\{p \mid \hat{H}(p) > 0 \}, \\
        \mathcal{P}^{fp} &=\{p \mid \hat{H}(p) = 0 \text{ and } H(p) > 0 \}, \\
    \end{aligned}
\end{equation}
where $\mathcal{P}^{pos}$ and $\mathcal{P}^{fp}$ is the set of positive and false-positive pixels, respectively.
Then we sample the ratio of $|\mathcal{P}^{pos}|$ and $|\mathcal{P}^{fp}|$ to be 1:3.
Here, the order of $\mathcal{P}^{fp}$ is sorted according to the distance between $\hat{H}$ and $H$ and then sampled in the top-k manner.

The final objective function $L$ is defined as,
\begin{equation}
    L = \frac{1}{S} \sum_{p \in \mathcal{P}}{M(p) \times (H(p) - \hat{H}(p))^2},
\end{equation}
where $\mathcal{P}$ is the set of sampled pixels from FPEM.

 \subsection{From heatmap to oriented bounding boxes} \label{map2box}
\begin{figure}[t]
    \centering
    \subfigure[]{           
        \includegraphics[width=0.22\linewidth]{ 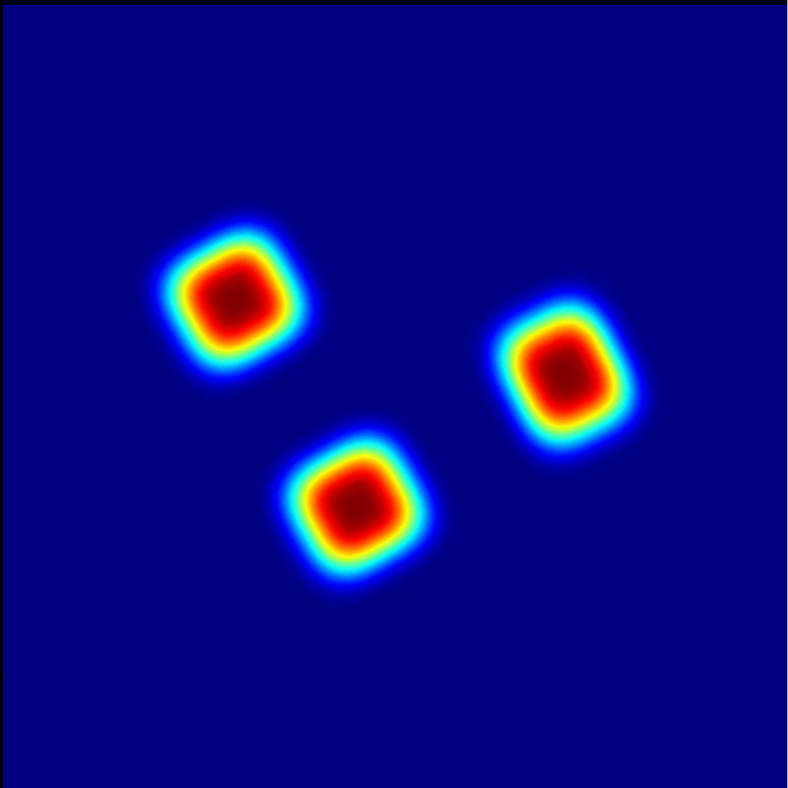} 
        \label{ori_post:1}          
    }
    \subfigure[]{           
        \includegraphics[width=0.22\linewidth]{ 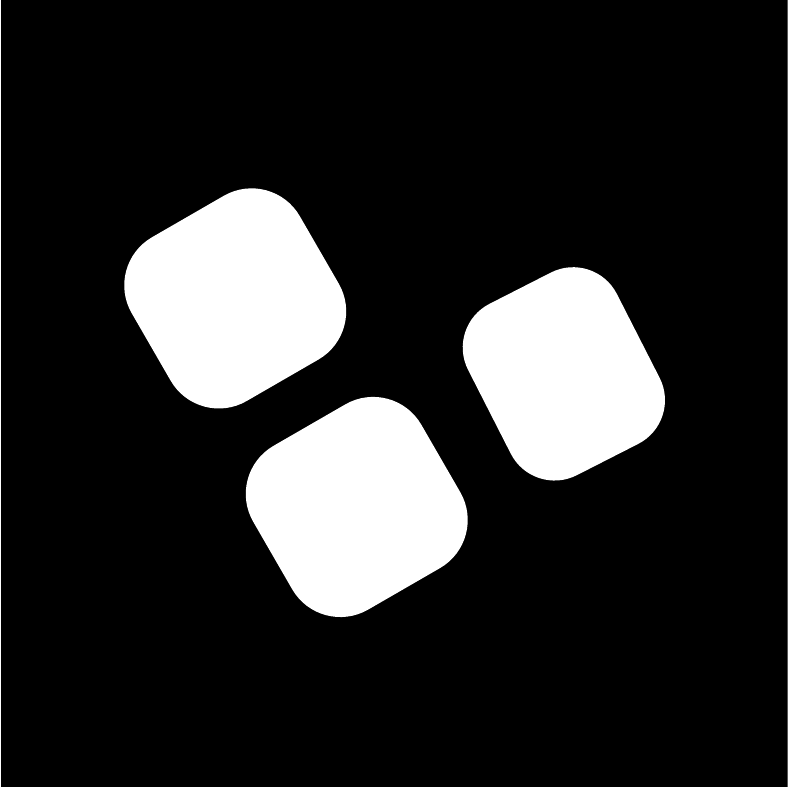} 
        \label{ori_post:2}          
    }
    \subfigure[]{           
        \includegraphics[width=0.22\linewidth]{ 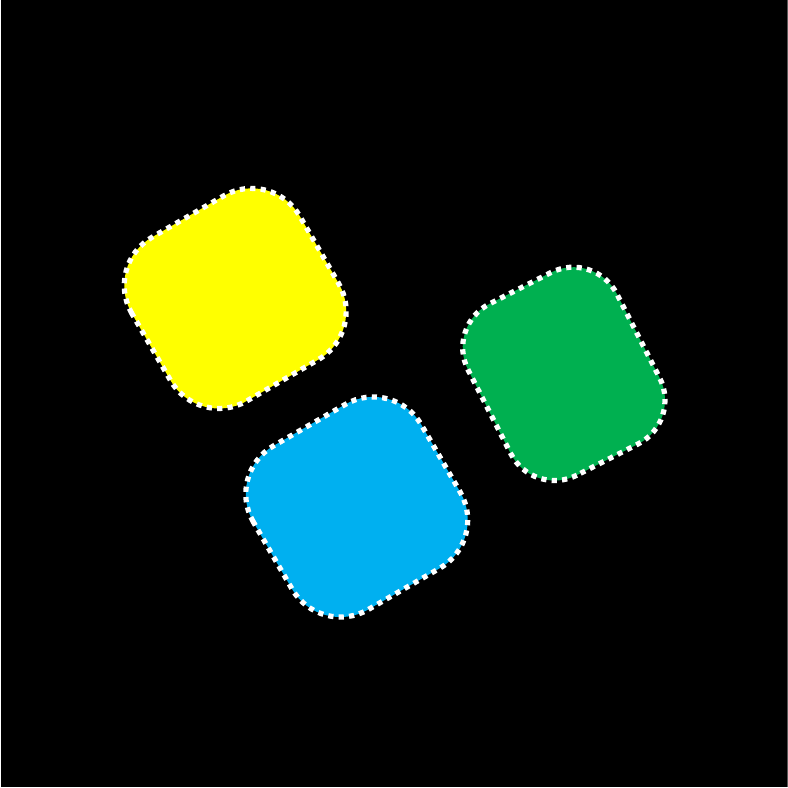} 
        \label{ori_post:3}          
    }
    \subfigure[]{           
        \includegraphics[width=0.22\linewidth]{ 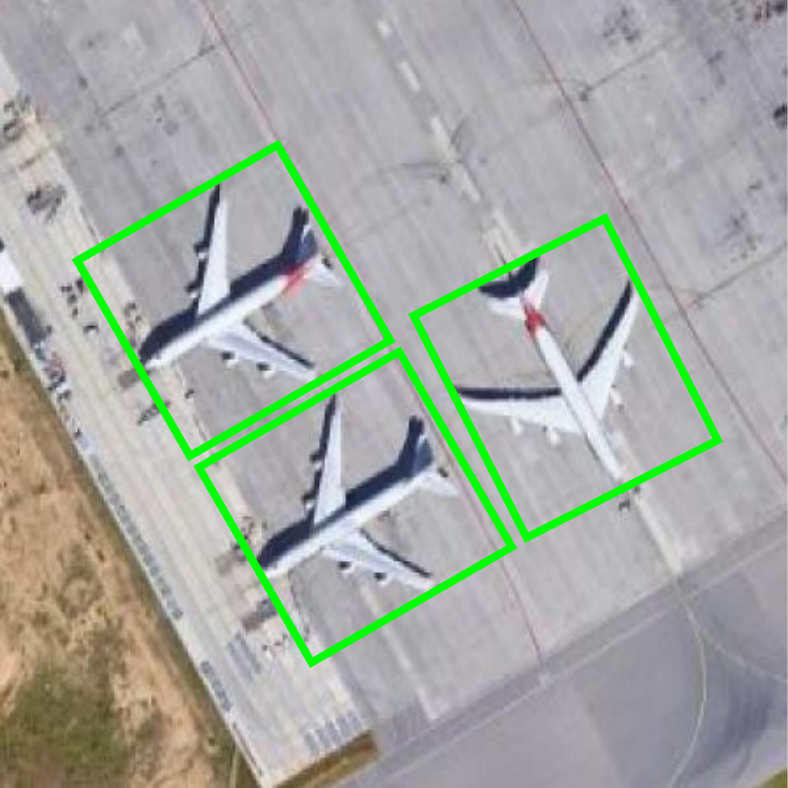} 
        \label{ori_post:4}          
    }
    \caption{ From heatmap to oriented bounding boxes. First, (a) from the heatmap, (b) we obtain binary heatmap using threshold $\tau$. (c) We label each kernel using the connected component labeling (CCL) algorithm and (d) extract the contour points and find the minimum-area rectangle for each kernel. }
    \label{ori_post}
\end{figure}

 We apply simple post-processing algorithms to extract the oriented bounding boxes from the heatmap, as illustrated in Figure ~\ref{ori_post}.
 First, we obtain a binary heatmap $S \in [0,1]^{\frac{H}{R} \times \frac{W}{R} \times C}$ from the output heatmap $H$ with a threshold $\tau$. 
 By increasing the $\tau$, we can separate the weakly-occluded objects.
 Second, through the connected component labeling (CCL)~\cite{he2009fast} algorithm, we allocate an ID of each Tricube kernel.
 Third, we obtain the rotated rectangular box with a minimum area containing each Tricube kernel using OpenCV~\cite{opencv_library} functions. 
 Last, since we drop some part of the Tricube kernel through the threshold $\tau$, we scale up the kernel back to its original size by multiplying a scale factor $s$.
 For the choice of $\tau$ and $s$, we adaptively set the $s$ according to the $\tau$ utilizing the distribution of the Tricube kernel, $i.e.,$ $y=(1-|x|^3)^\gamma$.
 Note that the distribution of the Tricube kernel is maintained even if the input resolution or the object size is changed, therefore, we can adaptively set $s=\frac{1}{1-(1-|\tau|^3)^{\gamma}}$, where $\gamma$ is set to 7 as described in \ref{method:kernel}.

%-------------------------------------------------------------------------
%-------------------------------------------------------------------------

%-------------------------------------------------------------------------
\begin{table*}[t]
  \centering
  \caption{
    Experiment results on DOTA dataset. MS and Flip denote the multi-scale and flip test time augmentation, respectively. PL-Plane, BD-Baseball Diamond, BR-Bridge, GTF-Ground Field Track, SV-Small Vehicle, LV-Large Vehicle, SH-Ship, TC-Tennis Court, BC-Basketball Court, ST-Storage Tank, SBF-Soccer Ball Field, RA-Roundabout, HA-Harbor, SP-Swimming Pool, and HC-Helicopter.
  }
  \label{tab:result_dota}
  \vspace{1mm}
  \begin{adjustbox}{max width=\linewidth}
  \begin{tabular}{c|c|cc|ccccccccccccccc|c}
    \hline
    Method & Backbone & MS & Flip & PL & BD & BR & GTF & SV & LV & SH & TC & BC & ST & SBF & RA & HA & SP & HC & mAP  \\
    
    \hline
        \textbf{two-stage} \\
    \hline
        R-DFPN~\cite{yang2018automatic}  & ResNet-101  & & & 80.92 & 65.82 & 33.77 & 58.94 & 55.77 & 50.94 & 54.78 & 90.33 & 66.34 & 68.66 & 48.73 & 51.76 & 55.10 & 51.32 & 35.88 & 57.94  \\
        RRPN~\cite{ma2018arbitrary}  &  ResNet-101 & & & 80.94 & 65.75 & 35.34 & 67.44 & 59.92 & 50.91 & 55.81 & 90.67 & 66.92 & 72.39 & 55.06 & 52.23 & 55.14 & 53.35 & 48.22 & 61.01  \\
        R$^2$CNN~\cite{jiang2017r2cnn}  & ResNet-101 & & & 88.52 & 71.20 & 31.66 & 59.30 & 51.85 & 56.19 & 57.25 & 90.81 & 72.84 & 67.38 & 56.69 & 52.84 & 53.08 & 51.94 & 53.58 & 60.67 \\
        ICN~\cite{azimi2018towards}  & ResNet-101 & & & 81.36 & 74.30 & 47.70 & 70.32 & 64.89 & 67.82 & 69.98 & 90.76 & 79.06 & 78.20 & 53.64 & 62.90 & 67.02 & 64.17 & 50.23 & 68.16  \\
        RoI Trans~\cite{ding2019learning}  &  ResNet-101 & \checkmark & & 88.64 & 78.52 & 43.44 & 75.92 & 68.81 & 73.68 & 83.59 & 90.74 & 77.27 & 81.46 & 58.39 & 53.54 & 62.83 & 58.93 & 47.67 & 69.56   \\
        SCRDet~\cite{yang2019scrdet}  &  ResNet-101 & & & 89.41 & 78.83 & 50.02 & 65.59 & 69.96 & 57.63 & 72.26 & 90.73 & 81.41 & 84.39 & 52.76 & 63.62 & 62.01 & 67.62 & 61.16 & 69.83   \\
        SCRDet~\cite{yang2019scrdet}  &  ResNet-101 & \checkmark & & 89.98 & 80.65 & 52.09 & 68.36 & 68.36 & 60.32 & 72.41 & 90.85 & \textbf{87.94} & \textbf{86.86} & \textbf{65.02} & 66.68 & 66.25 & 68.24 & 65.21 & 72.61   \\
        Gliding vertex~\cite{xu2020gliding}  &  ResNet-101 & \checkmark & & 89.64 & \textbf{85.00} & 52.26 & \textbf{77.34} & 73.01 & 73.14 & \textbf{86.82} & 90.74 & 79.02 & 86.81 & 59.55 & \textbf{70.91} & 72.94 & \textbf{70.86} & 57.32 & 75.02   \\
    \hline
        \textbf{one-stage} \\
    \hline
        FR-O~\cite{xia2018dota}  &  ResNet-101  &  &  & 79.42 & 77.13 & 17.70 & 64.05 & 53.30 & 38.02 & 37.16 & 89.41 & 69.64 & 59.28 & 50.30 & 52.91 & 47.89 & 47.40 & 46.30 & 54.13 \\
        RetinaNet~\cite{lin2017focal}  &  ResNet-50  & \checkmark &  & 88.87 & 74.46 & 40.11 & 58.03 & 63.10 & 50.61 & 63.63 & 90.89 & 77.91 & 76.38 & 48.26 & 55.85 & 50.67 & 60.23 & 34.23 & 62.22 \\
        R$^3$Det~\cite{yang2019r3det}  &  ResNet-101  & \checkmark &  & 89.54 & 81.99 & 48.46 & 62.52 & 70.48 & 74.29 & 77.54 & 90.80 & 81.39 & 83.54 & 61.97 & 59.82 & 65.44 & 67.46 & 60.05 & 71.69 \\
        R$^3$Det~\cite{yang2019r3det}  &  ResNet-152  & \checkmark &  & 89.49 & 81.17 & 50.53 & 66.10 & 70.92 & 78.66 & 78.21 & 90.81 & 85.26 & 84.23 & 61.81 & 63.77 & 68.16 & 69.83 & 67.17 & 73.74 \\
        RSDet~\cite{qian2019learning}  &  ResNet-101  & \checkmark &  & 89.80 & 82.90 & 48.60 & 65.20 & 69.50 & 70.10 & 70.20 & 90.50 & 85.60 & 83.40 & 62.50 & 63.90 & 65.60 & 67.20 & 68.00 & 72.20 \\
        RSDet~\cite{qian2019learning}  &  ResNet-152  & \checkmark &  & \textbf{90.10} & 82.00 & \textbf{53.80} & 68.50 & 70.20 & \textbf{78.70} & 73.60 & \textbf{91.20} & 87.10 & 84.70 & 64.30 & 68.20 & 66.10 & 69.30 & 63.70 & 74.10 \\
    \hline
        \textbf{anchor-free} \\
    \hline
        CenterNet~\cite{zhou2019objects}  &  Hourglass-104 &   &  & 89.02 & 69.71 & 37.62 & 63.42 & 65.23 & 63.74 & 77.28 & 90.51 & 79.24 & 77.93 & 44.83 & 54.64 & 55.93 & 61.11 & 45.71 & 65.04 \\
        CenterNet~\cite{zhou2019objects}  &  Hourglass-104 & \checkmark  &  & 89.56 & 79.83 & 43.80 & 66.54 & 65.58 & 66.09 & 83.11 & 90.72 & 83.72 & 84.30 & 55.62 & 58.71 & 62.48 & 68.33 & 50.77 & 69.95 \\
        DRN~\cite{pan2020dynamic}  &  Hourglass-104 &   &  & 88.91 & 80.22 & 43.52 & 63.35 & 73.48 & 70.69 & 84.94 & 90.14 & 83.85 & 84.11 & 50.12 & 58.41 & 67.62 & 68.60 & 52.50 & 70.70 \\
        DRN~\cite{pan2020dynamic}  &  Hourglass-104 & \checkmark  &  & 89.45 & 83.16 & 48.98 & 62.24 & 70.63 & 74.25 & 83.99 & 90.73 & 84.60 & 85.35 & 55.76 & 60.79 & 71.56 & 68.82 & 63.92 & 72.95 \\
        DRN~\cite{pan2020dynamic}  &  Hourglass-104 & \checkmark  & \checkmark & 89.71 & 82.34 & 47.22 & 64.10 & 76.22 & 74.43 & 85.84 & 90.57 & 86.18 & 84.89 & 57.65 & 61.93 & 69.30 & 69.63 & 58.48 & 73.23 \\
        TricubeNet (ours) &  Hourglass-104 & & & 87.51 & 73.62 & 43.21 & 63.67 & 76.97 & 72.97 & 84.36 & 89.21 & 83.59 & 84.60 & 47.29 & 61.77 & 73.36 & 68.74 & 69.40 & 72.17 \\
        TricubeNet (ours) &  Hourglass-104 & \checkmark & & 88.28 & 80.46 & 47.32 & 70.09 & 76.97 & 72.97 & 84.52 & 90.73 & 83.87 & 84.60 & 56.92 & 62.91 & 73.36 & 68.74 & 71.63 & 74.22 \\
        TricubeNet (ours) &  Hourglass-104 & \checkmark & \checkmark & 88.75 & 82.12 & 49.24 & 72.98 & \textbf{77.64} & 74.53 & 84.65 & 90.81 & 86.02 & 85.38 & 58.69 & 63.59 & \textbf{73.82} & 69.67 & \textbf{71.08} & \textbf{75.26} \\
    \hline
  \end{tabular}
  \end{adjustbox}
  \vspace{2mm}
\end{table*}
%-------------------------------------------------------------------------

%-------------------------------------------------------------------------
\begin{table}[t]
  \centering
  \caption{Evaluation results on SKU110K-R using the COCO-style metric.}
  \vspace{1mm}
  \label{tab:result_sku}
    \begin{adjustbox}{max width=\linewidth}
      \begin{tabular}{c|c|ccc}
        \hline
        Method & mAP & AP$_{50}$ & AP$_{75}$ & AR$_{300}$ \\
        \hline
        YOLOv3-Rotate~\cite{redmon2018yolov3}   &  49.1 & - & 51.1 & 58.2    \\
        CenterNet-4point~\cite{zhou2019objects} &  34.3 & - & 19.6 & 42.2    \\
        CenterNet~\cite{zhou2019objects}        &  54.7 & - & 61.1 & 62.2    \\
        DRN~\cite{pan2020dynamic}               &  55.9 & - & 63.1 & 63.3    \\
        
        \hline
        TricubeNet (ours) &  57.7 & 94.7 & 65.2 & 62.5    \\
        \hline
      \end{tabular}
    \end{adjustbox}
\end{table}
%-------------------------------------------------------------------------

 \section{Experiments}
 
 \subsection{Dataset}
 To demonstrate our effectiveness, we experiment on various practical datasets of weakly-occluded objects: aerial image, densely packed object image, and scene text image.
 
\noindent \textbf{DOTA} \cite{xia2018dota}
 is the largest dataset for oriented object detection in aerial images; it consists of 2806 images with sizes ranging from 800$\times$800 to 4000$\times$4000. The number of training, validation, and testing images are 1411, 458, and 937, respectively. It contains 15 categories of objects and 188,282 instances with a wide variety of scales, orientations, and shapes.
Following other conventions~\cite{ding2019learning, pan2020dynamic}, we resize the image and crop a series of 1024$\times$1024 patches from the original images with a stride of 824.
For training, we resize the images at three scales (0.75, 1.0, and 1.5).
For single-scale testing and multi-scale testing, we resize the images at one scale (1.0) and three scales (0.75, 1.0, and 1.5), respectively.
The performance on the test set is measured on the official DOTA evaluation server\footnote{\url{https://captain-whu.github.io/DOTA}}.

\noindent \textbf{SKU110K-R~\cite{pan2020dynamic}}
is an extended dataset of SKU110K~\cite{goldman2019precise} for the densely packed oriented object detection.
It contains thousands of supermarket store images with various scales from 1840$\times$1840 to 4320$\times$4320, viewing angles, lighting conditions, and noise levels.
Each image contains an average of 154 tightly packed objects, up to 718 objects.
The rotation data augmentation with six angles (-45$^{\circ}$, -30$^{\circ}$, -15$^{\circ}$, 15$^{\circ}$, 30$^{\circ}$, and 45$^{\circ}$) is performed to original images.
After the augmentation, the number of training, validation, and testing images are 57,533, 4,116, and 20,587, respectively.

\noindent \textbf{MSRA-TD500} \cite{yao2012detecting}
is a dataset for multi-lingual, long, and oriented text detection in both indoors and outdoors natural images.
The images contain English and Chinese scripts and each text is labeled by a rotated rectangle.
It consists of 300 training images and 200 testing images.

\noindent \textbf{ICDAR2015} \cite{karatzas2015icdar}
is proposed in Robust Reading Competition for incidental scene text detection.
There are 1000 training images and 500 testing images.
Each text is annotated as word level with a quadrangle of four vertexes.

 \subsection{Training details} \label{train_detail}
 For SKU110K-R, MSRA-TD500, and ICDAR2015 datasets, we set the input resolution to 800$\times$800 and apply random cropping, random rotating, color jittering for data augmentation.
 For DOTA dataset, we set the input resolution to 1024$\times$1024 and apply random rotating and color jittering data augmentation.
 Adam~\cite{kingma2014adam} is used as the optimizer and the learning rate is set to $2.5e$-$4$ for all datasets.
 For DOTA dataset, we train the model with 140 epochs and drop the learning rate by a factor 10 at 90 and 120 epochs.
 For SKU110K-R dataset, we train the model with 20 epochs without the learning rate decay.
 Following existing text detection approaches~\cite{deng2018pixellink, choi2019gaussian}, we pre-train the TricubeNet on SynthText~\cite{gupta2016synthetic} dataset with 3 epochs and finetuned on MSRA-TD500 and ICDAR2015 datasets. 
 For MSRA-TD500 and ICDAR2015 datasets, we train the model with 180 epochs and drop the learning rate by a factor 10 at 120 and 160 epochs.

 \subsection{Testing details} \label{test_detail}
 We extract oriented bounding boxes using the post-processing algorithm described in Section~\ref{map2box}.
 For DOTA dataset, we apply two kinds of test time augmentation: multi-scale testing and flip augmentation.
 For multi-scale testing, three scales (0.75, 1.0, and 1.5) and Soft-NMS for oriented bounding boxes are used.
 For the flip augmentation, we average output heatmaps.
 For the evaluation metric of DOTA, MSRA-TD500, and ICDAR2015, we adopt the mean average precision (mAP) of the 0.5 polygon IoU threshold.
 For the SKU110K-R dataset, we adopt COCO-style~\cite{lin2014microsoft} evaluation method for oriented boxes: mAP at IoU=0.5:0.05:0.95, average precision AP$_{75}$ at IoU of 0.75, and average recall AR$_{300}$ at IoU=0.5:0.05:0.95.

%-------------------------------------------------------------------------
\begin{table}[t]
  \centering
  \caption{Evaluation on MSRA-TD500 and ICDAR2015 testset. $*$ denotes for multi-scale test. R, P, and H denote recall, precision, and H-mean respectively.}
  \vspace{1mm}
  \label{tab:result_text}
    \begin{adjustbox}{max width=\linewidth}
      \begin{tabular}{c|ccc|ccc}
        \hline
        & \multicolumn{3}{c|}{MSRA-TD500} & \multicolumn{3}{c}{ICDAR 2015}  \\
        Method & R & P & H & R & P & H  \\
        \hline
        \textbf{two-stage} \\
        \hline
        Wang \textit{et al.}~\cite{wang2019arbitrary}  &  82.1  & 85.2 & 83.6 & \textbf{86.0} & 89.2 & \textbf{87.6}  \\
        Gliding vertex~\cite{xu2020gliding}        &  \textbf{84.3}  & 88.8 & \textbf{86.5} & -    & -    & -     \\
        \hline
        \textbf{one-stage} \\
        \hline
        SegLink~\cite{shi2017detecting}               &  70.0  & 86.0 & 77.0 & 76.8 & 73.1 & 75.0  \\
        RRD*~\cite{liao2018rotation}                  &  73.0  & 87.0 & 79.0 & 80.0 & 88.0 & 83.8  \\
        Lyu \textit{et al.}*~\cite{lyu2018multi}  &  76.2  & 87.6 & 81.5 & 79.7 & 89.5 & 84.3  \\
        Direct*~\cite{he2018multi}               &  81.0  & \textbf{91.0} & 86.0 & 80.0 & 85.0 & 82.0  \\
        \hline
        \textbf{anchor-free} \\
        \hline
        Zhang \textit{et al.}~\cite{zhang2016multi} &  67.0  & 83.0 & 74.0 & 43.0 & 71.0 & 54.0  \\
        EAST*~\cite{zhou2017east}                   &  67.4  & 87.3 & 76.1 & 78.3 & 83.3 & 80.7  \\
        TextSnake~\cite{long2018textsnake}             &  73.9  & 83.2 & 78.3 & 80.4 & 84.9 & 82.6  \\
        PixelLink*~\cite{deng2018pixellink}            &  73.2  & 83.0 & 77.8 & 82.0 & 85.5 & 83.7  \\
        CRAFT~\cite{baek2019character}                 &  78.2  & 88.2 & 82.9 & 84.3 & 89.8 & 86.9  \\
        TricubeNet (ours)  &  80.8  & 90.4 & 85.3 & 75.2 & \textbf{90.1} & 82.0  \\
        \hline
      \end{tabular}
    \end{adjustbox}
    \vspace{2mm}
\end{table}
%-------------------------------------------------------------------------

 \subsection{State-of-the-art comparisons}
 We compare the performance of TricubeNet with the state-of-the-art methods on the test set of each dataset: Table \ref{tab:result_dota} for DOTA, Table \ref{tab:result_sku} for SKU110K-R, Table \ref{tab:result_text} for MSRA-TD500 and ICDAR2015.
 For the comparison, we apply two-steps heatmap cascade refinement.
 The anchor-based two-stage and one-stage detectors require a huge number of anchor boxes and multiple loss functions for the box offsets regression; it demands huge computational complexity and careful hyperparameter tuning.
 Meanwhile, anchor-free detectors can alleviate the above problems concerned with the anchor box by eliminating the anchor box in the network design, but their performance is slightly inferior to that of anchor-based detectors.
 However, our TricubeNet achieves highly competitive performance in all datasets despite its anchor-free setting.
 Especially, our outstanding performance in small vehicles (SV) shows that TricubeNet can accurately detect small objects without anchor box tuning, and we can obtain well-arranged oriented boxes as shown in Figure \ref{fig:qualitative_dota}.
 In addition, the sufficiently high performance of 94.7\% $AP_{50}$ on SKU110K-R shows that TricubeNet is effective in detecting densely packed objects.
 When applying the same test augmentation, Gliding vertex~\cite{xu2020gliding}, one of the two-stage detectors, shows the state-of-the-art performance (75.0\% $v.s.$ 74.2\%) as in Table \ref{tab:result_dota}, we argue that our efficiency is higher than them because we exploit the advantages of the one-stage anchor-free setting and simply solve the loss discontinuity problem.

 \subsection{Ablation study} \label{section_ablation}
 
 To analyze the effectiveness of each component of TricubeNet, we conduct an ablation study.
 For the evaluation, we experiment on DOTA validation set following all parameter settings as given in Section~\ref{train_detail}.

 %-------------------------------------------------------------------------
\begin{table}[t]
    \centering
    \caption{
        The effect of each kernel representing an object.
      }
    \vspace{1mm}
    \label{tab:ablation_kernel}
    \begin{adjustbox}{max width=\linewidth}
    \begin{tabular}{c|cccc}
        \hline
        kerenl & Tricube & Gaussian & Effective Rect & Binary Rect \\
        \hline
        mAP & 75.26 & 72.12  &  69.31    &   58.52 \\
        \hline
      \end{tabular}
    \end{adjustbox}
\end{table}
%-------------------------------------------------------------------------

 %-------------------------------------------------------------------------
 \begin{table}[t]
    \centering
    \caption{
        Ablation study for the proposed techniques: size-weight mask (SWM), false-positive example mining (FPEM), multi-angle convolution (MAC) module, one-step heatmap refinement (Cascade 1), and two-step heatmap refinement (Cascade 2).
      }
    \vspace{1mm}
    \label{tab:ablation_comp}
    \begin{adjustbox}{max width=\linewidth}
      \begin{tabular}{c|ccc|ccc|cc}
        \hline
                     & $AP$   & $AP_{50}$ & $AP_{75}$ & $AP_{S}$ &  $AP_{M}$ &  $AP_{L}$ & Params & GFlops \\
        \hline
        Baseline     & 30.4 & 61.6    &   25.2  &  16.3  &   32.4  &   37.6  &  188M  &  1015G \\
        + SWM        & 31.1 & 64.0    &   25.6  &  19.2  &   33.0  &   36.0  &  +0M   &  +0G \\
        + FPEM       & 32.3 & 66.8    &   25.2  &  20.5  &   35.4  &   37.6  &  +0M   &  +0G \\
        + MAC        & 33.6 & 67.7    &   28.7  &  20.2  &   35.9  &   39.8  &  +0.4M &  +28G \\
        + Cascade 1  & 35.3 & 69.6    &   30.3  &  24.1  &   37.2  &   40.1  &  +1.2M &  +85G \\
        + Cascade 2  & 35.6 & 70.1    &   30.9  &  24.6  &   38.5  &   40.3  &  +2.1M &  +143G \\
        
        \hline
      \end{tabular}
    \end{adjustbox}
    \vspace{2mm}
\end{table}
%-------------------------------------------------------------------------

 %-------------------------------------------------------------------------
 \begin{table}[t]
    \centering
    \caption{
      The analysis of each method for extracting rotation-invariant features.
      }
    \vspace{1mm}
    \label{tab:ablation_conv}
    \begin{adjustbox}{max width=\linewidth}
      \begin{tabular}{c|ccc|ccc|cc}
        \hline
                     & $AP$   & $AP_{50}$ & $AP_{75}$ & Params & GFlops \\
        \hline
        CNN                          &  32.3  &  66.8    &  25.2  &  +0.00M &  +00G  \\
        DCN~\cite{zhu2019deformable} &  32.9  &  67.0    &  28.1  &  +0.55M &  +40G  \\
        RCL~\cite{pan2020dynamic}    &  33.4  &  67.6    &  28.5  &  +0.49M &  +34G  \\
        MAC (ours)                   &  33.6  &  67.7    &  28.7  &  +0.43M &  +28G  \\
        \hline
      \end{tabular}
    \end{adjustbox}
\end{table}
%-------------------------------------------------------------------------

 First, we investigate which 2D kernel is more effective in representing the object.
 For the comparison, we choose one of four different types of 2D kernels: 2D Tricube kernel, 2D Gaussian kernel, binary rectangle, effective rectangle.
 The effective rectangle is a shrunk binary rectangle, which is employed in FSAF~\cite{zhu2019feature}, and we use a binary rectangle shrunk by 40\% as the effective rectangle.
 Table \ref{tab:ablation_kernel} shows the DOTA validation score of each kernel.
 The effective rectangle achieves higher performance than the binary rectangle because the binary rectangle often fails to separate weakly-occluded objects.
 Although the Gaussian kernel has a limitation in representing the angle of the object as discussed in Section \ref{method:kernel}, the Gaussian kernel shows a 2.81\% higher performance than the effective rectangle; this result demonstrates the advantage of the kernel-based object representation.
 Since the Tricube kernel can solve the limitation of the Gaussian kernel while taking advantage of the kernel-based object representation, the Tricube kernel is the most suitable choice for the oriented object and achieves the highest performance among the four types of kernels.
 
 Second, we evaluate the effect of proposed techniques, $i.e.,$ size-weight mask (SWM), false-positive example mining (FPEM), multi-angle convolution (MAC) module, and heatmap cascade refinement.
 For a more precise analysis, we measure the performance using COCO-style evaluation metric as in Table \ref{tab:ablation_comp}.
 $AP_{S}$, $AP_{M}$, and $AP_{L}$ denote the AP when the object size is smaller than 32~$\times$~32, between 32~$\times$~32 and 96~$\times$~96, and larger than 96~$\times$~96, respectively.
 In addition, we report the number of parameters and GFlops.
 We set the baseline model as the hourglass-104 network trained using MSE loss without any proposed techniques and evaluate the effect of each technique.
 When applying both SWM and FPEM, $AP$ and $AP_{S}$ are considerably improved by 1.9\% and 4.2\%, respectively; it clearly shows that SWM and FPEM help in solving size-imbalance and class-imbalance problems.
 The MAC module, which is designed for extracting the rotation-invariant feature, yields 1.3\% performance improvement.
 One-step heatmap refinement further improves the performance by 2.3\%.
 However, two-step heatmap refinement only yields 0.3\% improvement; we conclude that two-step heatmap refinement is enough to produce a high-quality heatmap.
 
 Last, in Table \ref{tab:ablation_conv}, we evaluate our MAC module by comparing with a conventional convolution (CNN), deformable convolution~\cite{zhu2019deformable} (DCN), and rotation convolution layer~\cite{pan2020dynamic} (RCL).
 Compared to CNN and DCN, our MAC module can properly extract rotation-invariant features, so the performance is improved by 1.3\% and 0.7\%, respectively.
 Since both MAC and RCL are designed to extract rotation-invariant features, our MAC module shows a similar performance improvement to the RCL.
 However, the RCL requires heavier computations and about 60\% more GPU memory usage than our MAC during the training.
 From the above result, we prove that the proposed MAC module is more effective and efficient in our TricubeNet, which is a fully-heatmap-based approach.

 \subsection{Qualitative Results} \label{section_qualitative}
\begin{figure}[t]
    \centering
    \subfigure[DOTA]{           
        \includegraphics[width=\linewidth]{ 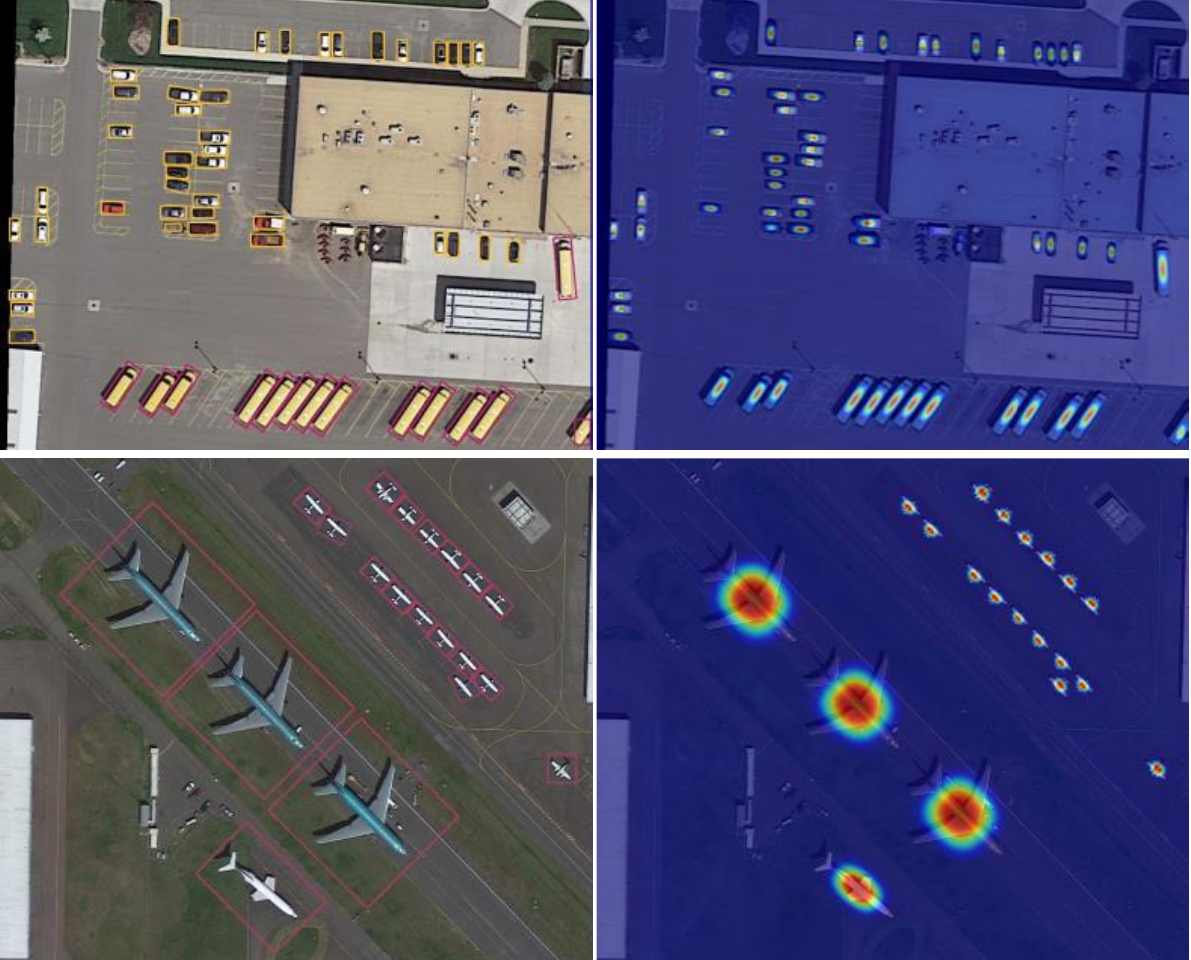} 
        \label{fig:qualitative_dota}       
    }
    \vspace{-0.75mm}
    \subfigure[MSRA-TD500]{           
        \includegraphics[width=\linewidth]{ 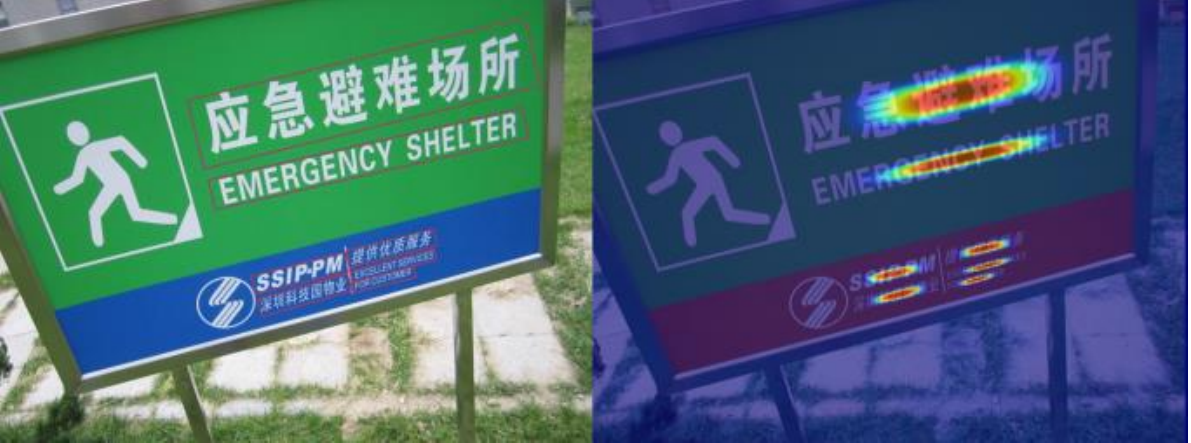} 
        \label{fig:qualitative_msra}      
    }
    \vspace{-0.75mm}
    \subfigure[ICDAR 2015]{           
        \includegraphics[width=\linewidth]{ 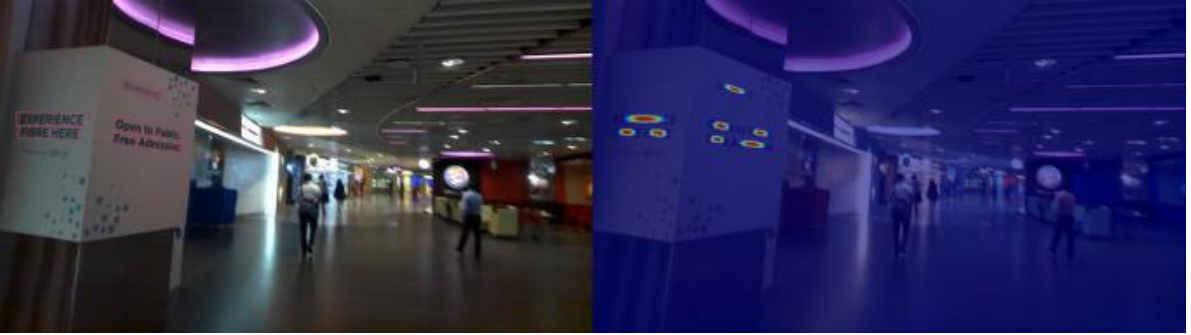} 
        \label{fig:qualitative_icdar}     
    }
    \vspace{-0.75mm}
    \subfigure[SKU110K-R]{           
        \includegraphics[width=\linewidth]{ 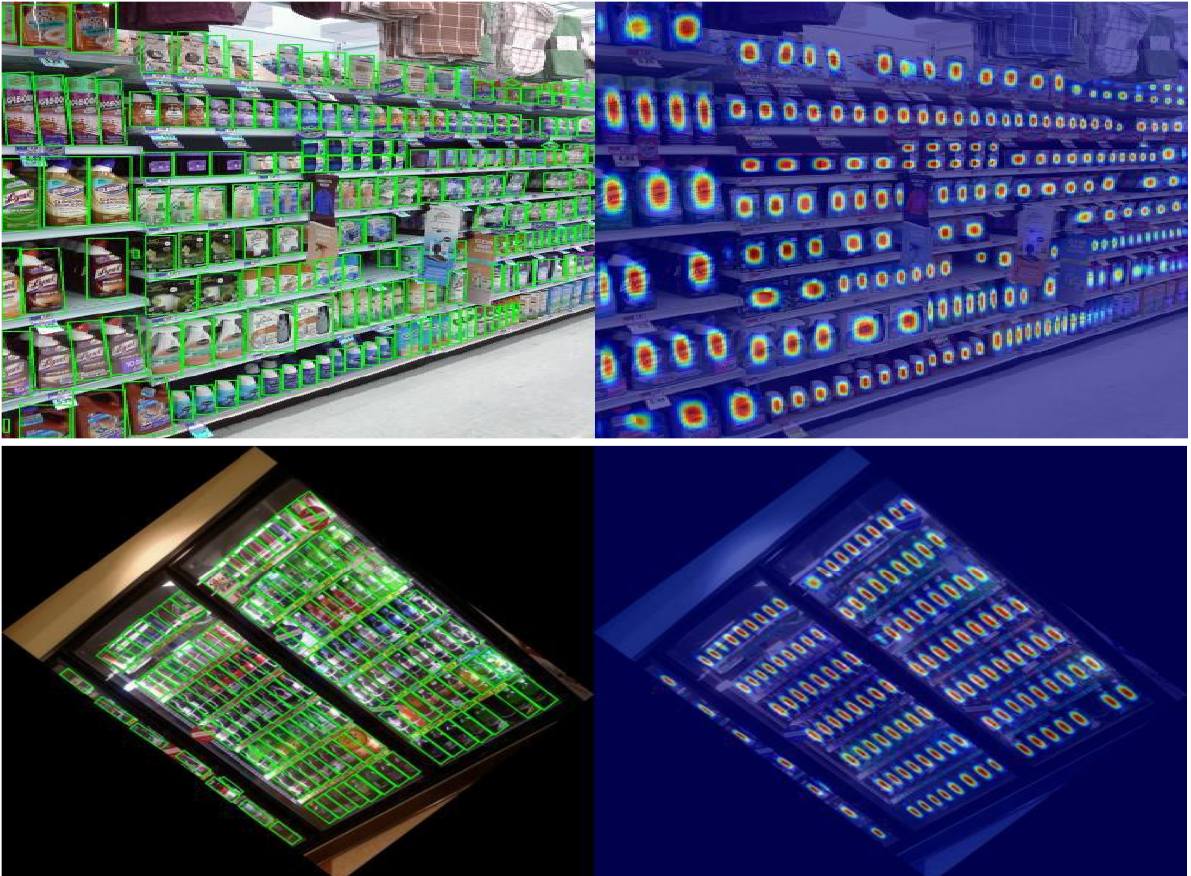} 
        \label{fig:qualitative_sku}
    }
    %\vspace{0.1mm}
    \caption{ Qualitative results of TricubeNet.}
    \label{fig:qualitative}
\end{figure}

 We collect some qualitative results on each dataset: Figure \ref{fig:qualitative_dota} for DOTA, Figure \ref{fig:qualitative_msra} for MSRA-TD500, Figure \ref{fig:qualitative_icdar} for ICDAR 2015, and Figure \ref{fig:qualitative_sku} for SKU110K-R.
 From the results, we can notice that our TricubeNet handles the challenging factors of oriented object detection.
 Specifically, as in Figure \ref{fig:qualitative_dota}, TricubeNet can properly detect objects of various sizes within the multi-category classification problem;
 as in Figure \ref{fig:qualitative_icdar}, TricubeNet can accurately detect the rotated text despite the complex background;
 as in Figure \ref{fig:qualitative_sku}, TricubeNet is outstanding in the detection of a huge number of densely arranged oriented objects.
 Furthermore, we can identify that the consistency of the rotated bounding boxes detecting the arranged objects is very high.

%-------------------------------------------------------------------------

 \section{Conclusion and Future Work}
   We present a novel approach for oriented object detection, named TricubeNet.
   Our main concept is that we localize objects using visual cues ($i.e.,$ heatmap) instead of box offsets regression.
   We represent each object as a 2D Tricube kernel and extract bounding boxes using simple image-processing algorithms.
   Unlike anchor-based approaches, TricubeNet is able to obtain well-arranged oriented boxes from visual cues, solve the angle discontinuity problem by taking away the angle regression, and save the computational complexity due to our anchor-free modeling.
   Additionally, for the better fit in oriented object detection, we propose some effective techniques: SWM for the size-invariant loss function; FPEM for balancing between foreground and background pixels and reducing false-positive detections; MAC module to extract the rotation-invariant features; heatmap cascade refinement for progressively refine the output heatmap.
   The extensive experimental results on various datasets show that our TricubeNet is considerably effective and efficient for oriented object detection.
   
   For future work, our failure cases are should be addressed.
   TricubeNet is quite effective to detect weakly-occluded oriented objects, however, it has trouble detecting heavily occluded objects.
   Also, the extracted box offsets from our post-processing algorithm are integer type, which weakens the precise detection of tiny objects.
   Addressing these limitations would give a great improvement.

\section*{Acknowledgement}
This research was supported in part by Autonomous Driving Center, R\&D Division, Hyundai Motor Company when the authors worked at KAIST.

\nocite{redmon2017yolo9000}
\nocite{yang2018position}
\nocite{zhou2020arbitrary}
\nocite{sun2019deep}

{\small
\bibliographystyle{ieee_fullname}
\bibliography{ms}
}

\end{document}